\documentclass[10pt]{article} 
\usepackage[preprint]{tmlr}

\usepackage{amsmath,amsfonts,bm}

\def\eqref#1{equation~\ref{#1}}

\def\1{\bm{1}}

\DeclareMathAlphabet{\mathsfit}{\encodingdefault}{\sfdefault}{m}{sl}
\SetMathAlphabet{\mathsfit}{bold}{\encodingdefault}{\sfdefault}{bx}{n}

\usepackage{hyperref}
\usepackage{url}
\usepackage{graphicx}           
\usepackage{algorithm}
\usepackage{algpseudocode}
\usepackage{multirow}
\usepackage{booktabs}
\usepackage{subcaption}

\title{Consistency Trajectory Planning: High-Quality and Efficient Trajectory Optimization for Offline Model-Based Reinforcement Learning}

\author{\name Guanquan Wang \email guanquan-wang@g.ecc.u-tokyo.ac.jp\\
      \addr Department of Information and Communication Engineering\\
     The University of Tokyo
      \AND
      \name Takuya Hiraoka \email takuya-h1@nec.com \\
      \addr NEC Corporation, Tokyo, Japan
      \AND
      \name  Yoshimasa Tsuruoka \email yoshimasa-tsuruoka@g.ecc.u-tokyo.ac.jp\\
      \addr Department of Information and Communication Engineering \\
      The University of Tokyo}

\def\month{12}  
\def\year{2024} 
\def\openreview{\url{https://openreview.net/forum?id=TuACCzfty3}} 

\begin{document}

\maketitle
\begin{abstract}

This paper introduces Consistency Trajectory Planning (CTP), a novel offline model-based reinforcement learning method that leverages the recently proposed Consistency Trajectory Model (CTM) for efficient trajectory optimization. While prior work applying diffusion models to planning has demonstrated strong performance, it often suffers from high computational costs due to iterative sampling procedures. CTP supports fast, single-step trajectory generation without significant degradation in policy quality. We evaluate CTP on the D4RL benchmark and show that it consistently outperforms existing diffusion-based planning methods in long-horizon, goal-conditioned tasks. Notably, CTP achieves higher normalized returns while using significantly fewer denoising steps.  
In particular, CTP achieves comparable performance with over $120\times$ speedup in inference time, demonstrating its practicality and effectiveness for high-performance, low-latency offline planning.

\end{abstract}

\section{Introduction}
\label{sec:intro}
Recent advances in generative models have significantly impacted offline reinforcement learning (RL), particularly in trajectory planning tasks where agents must learn optimal behavior from fixed datasets. Among these models, diffusion-based methods have emerged as powerful tools for modeling complex, multimodal trajectory distributions. Notable examples include Diffuser \citep{janner2022planning}, which leverages score-based diffusion models with classifier guidance to generate high-return trajectories, and Decision Diffuser \citep{ajay2022conditional}, which employs classifier-free guidance by directly conditioning on returns during sampling. While effective, both approaches rely on iterative sampling procedures during inference—often requiring dozens or even hundreds of function evaluations—posing severe limitations for real-time decision-making.

To address the inefficiencies inherent in iterative diffusion-based sampling, recent work has explored distilled models such as Consistency Model (CM) \citep{song2023consistency}, which bypasses the reverse diffusion process by directly learning the mapping from noise to data \citep{wang2022diffusion,kang2024efficient}. In the context of model-free RL, CM has demonstrated substantial speedup with only marginal performance degradation \citep{ding2023consistency, wangplanning}. However, a critical limitation remains: CM lacks a principled mechanism to trade off between sampling speed and sample quality. This stems from the nature of the distillation process, where the consistency function is trained to project arbitrary intermediate states along the ODE trajectory back to the clean data. As a result, in practice, the multistep sampling procedure of CM for improved sample quality alternates between denoising and injecting noise. This iterative refinement, however, accumulates errors particularly as the number of function evaluations increases.

Building upon this observation, Consistency Trajectory Model (CTM) has recently been proposed as a generalization of both score-based and consistency-based models \citep{kim2023consistency}. CTM enables anytime-to-anytime transitions along the probability flow ODE, supporting flexible and efficient generation through both short-step and long-jump sampling. CTM retains access to the score function while allowing diverse training losses such as denoising score matching and adversarial losses, ultimately improving expressiveness and generalization. A recent concurrent study \citep{duan2025accelerating} also investigates the use of CTM for improving inference efficiency in offline RL. The algorithm operates in a \textit{model-free} setting and primarily focuses on policy learning from demonstration data. Although this algorithm accelerates diffusion-based models for decision-making tasks, it offers limited improvements in long-horizon tasks.

Motivated by the strengths of CTM and the need for efficient planning in offline RL, we propose Consistency Trajectory Planning (CTP), a novel offline model-based RL algorithm that integrates CTM into the trajectory optimization process (Section \ref{sec:plan_consistency}). CTP inherits the speed and flexibility of CTM, allowing planners to efficiently navigate the trade-off between planning speed and return quality. Unlike previous score-based planners requiring classifier guidance or iterative refinements, our method enables fast one-step sampling while retaining controllability and sample diversity. Furthermore, CTM's access to score information enables conditional planning and return-conditioning without the need for learned Q-functions, which faces challenges due to overestimated Q-values for out-of-distribution actions \citep{kumar2020conservative,levine2020offline}.

We evaluate CTP on D4RL benchmark tasks \citep{fu2020d4rl} (Section \ref{sec:experiment}). Across multiple tasks, CTP consistently matches or outperforms prior diffusion-based planners and consistency policies while achieving significant improvements in inference speed—making it well-suited for real-time or high-frequency control applications.

\section{Related Work}
\label{related_work}

\subsection{Diffusion Models}
\label{sec:diffusion_models}

Diffusion models have emerged as a significant advancement in generative models, particularly noted for their capability in producing high-fidelity images and text \citep{saharia2022photorealistic, nichol2021improved}. By learning to reverse a process where noise is progressively added to the data \citep{sohl2015deep,ho2020denoising}, Diffusion models formulate the data sampling process as an iterative denoising procedure. 

An alternative perspective on this denoising process involves parameterizing the gradients of the data distribution to achieve the score matching objective, as clarified by \citet{hyvarinen2005estimation}. This interpretation situates diffusion models within the broader category of Energy-Based Models, a classification supported by the contributions of \citet{du2019implicit, nijkamp2019learning} and \citet{grathwohl2020learning}.

To facilitate the generation of images conditioned on additional information, such as text, prior work \citep{nichol2021improved} employed a classifier to guide sampling. However, more recent work by \citet{ho2022classifier} has introduced classifier-free guidance, which leverages gradients from an implicit classifier derived from the differences in score functions between conditional and unconditional diffusion models. This method has demonstrated better performance in enhancing the quality of conditionally generated samples compared to traditional classifier guidance techniques. 

\subsection{Diffusion Models in Reinforcement Learning}
\label{sec:DM_in_RL}

In the field of RL, diffusion models have been utilized as a flexible tool for dataset augmentation. For example, the SynthER framework \citep{lu2024synthetic} leverages unguided diffusion models to augment both offline and online RL datasets, which are subsequently employed by model-free off-policy algorithms. Although this approach improves performance, it faces challenges due to the distributional shift that arises when approximating the behavior distribution with unguided diffusion. Similarly, MTDiff \citep{he2024diffusion} applies unguided data generation techniques in multitask environments, expanding the utility of diffusion models in complex settings.

Moreover, diffusion models have been adapted for training world models in RL. For instance, \citet{alonso2023diffusion} utilize diffusion techniques to train world models capable of accurately predicting future observations. However, this method does not account for entire trajectories, resulting in error accumulation and a lack of policy guidance. To address these limitations, \citet{rigter2023world} incorporate policy guidance into a diffusion-based world model for online RL, thereby improving its effectiveness. In the context of offline RL, \citet{jackson2024policy} provide a theoretical framework that elucidates how policy guidance influences the trajectory distribution, offering a foundation for further research.

Diffusion models have also been employed for policy representation in RL, particularly for capturing the multi-modal distributions present in offline datasets. A notable example is Diffusion-QL \citep{wang2022diffusion}, which integrates diffusion models into both Q-learning and Behavior Cloning frameworks for policy representation. However, Diffusion-QL exhibits computational inefficiency due to the necessity of processing the entire diffusion process both forward and backward during training. To mitigate this issue, \citet{kang2024efficient} introduce action approximation, which eliminates the need for denoising during the training phase.

Additionally, recent studies have explored the application of diffusion models in human behavior imitation learning and trajectory generation for offline RL. Specifically, Diffuser \citep{janner2022planning} generates trajectories and applies guidance mechanisms to steer them toward high returns or specific goals. After producing a whole sequence, it executes only the first action and then re-plans using a receding horizon strategy. Similarly, Decision Diffuser \citep{ajay2022conditional} also operates at the trajectory level, but leverages conditional inputs—such as desired rewards or goals—for guidance, eliminating the need for an explicit reward model.

Despite the promising results achieved by these diffusion-based methods, they generally suffer from substantial computational overhead due to the requirement of iterative sampling. Both Diffuser and Decision Diffuser rely on generating entire action trajectories through multiple denoising steps, which are computationally expensive at inference time. Moreover, their reliance on repeated re-planning further exacerbates inefficiency, especially in long-horizon tasks. While methods such as Diffusion-QL and its variants attempt to address training-time costs, they do not fundamentally eliminate the need for sequential sampling during test-time planning.

\textbf{Positioning of our work.} Motivated by these limitations, our work introduces a more efficient alternative based on the recently proposed CTM, which supports single-step trajectory generation. By integrating CTM into the trajectory optimization process, our method circumvents the need for iterative denoising and enables fast, one-step planning with minimal performance degradation. In contrast to prior diffusion-based planners, CTP achieves a favorable trade-off between sample quality and computational cost, making it particularly well-suited for time-sensitive offline RL applications.

\section{Preliminary}
\label{sec:preliminary}
\subsection{Reinforcement Learning Problem Setting}
\label{sec:problem_setting}
RL has long been established as a powerful framework for solving sequential decision-making problems, typically formalized as a Markov Decision Process defined by the tuple: $M=\left \{ S,A,P,R,\gamma ,d_{0}  \right \}$, where $S$ is the state space, $A$ is the action space, $P: S\times A \rightarrow S$ is the transition function, $R: S\times A\times S\rightarrow \mathbb{R}$ is the reward function, $\gamma \in [0,1)$ is the discount factor, and $d_{0}$ is the initial state distribution. 
The objective of RL is to learn a policy that produces a sequence of actions $a_{0:k_{end}}^*$ which maximizes the expected cumulative discounted return, given by $\mathbb{E} \left [  {\textstyle \sum_{k=0}^{k_{end} }}\gamma^k r\left (s_{k},a_{k}  \right )    \right ]$, where $k_{end}$ is the index of final time step in a trajectory. 

\subsection{Consistency Models}
\label{sec:consistency_models}
Diffusion-based generative models synthesize data by progressively corrupting it with Gaussian noise and then applying a learned denoising process to recover samples. This process can be interpreted through a continuous-time formulation, where \citet{song2020score} propose modeling sample dynamics using a stochastic differential equation that tracks the evolution of noisy data over time. In its deterministic counterpart, data generation follows a probability flow ordinary differential equation (ODE) of the form: $\frac{\mathrm{d} \mathbf{x}_{t}}{\mathrm{d} t } =-t\nabla \log_{t}{p_{t}( \mathbf{x}) }$, where $p_t(\mathbf{x})$ represents the marginal distribution obtained by convolving the clean data distribution $p_{\text{data}}(\mathbf{x})$ with Gaussian noise of variance $t^2$. The denoising process traces this ODE in reverse, starting from a sample drawn from $\mathcal{N}(0, t_N^2\mathbf{I})$ and integrating back to time $\epsilon > 0$, a small constant introduced for numerical stability near the initial condition.

This reverse integration requires numerous fine-grained steps, making sampling computationally expensive. To overcome this issue, CM has been introduced as an efficient alternative that replaces the sequential denoising trajectory with a single-shot approximation. Instead of modeling transitions between adjacent time steps, the consistency framework learns a parameterized consistency function $f_\theta(\mathbf{x}_t, t) \approx \mathbf{x}_\epsilon$ that maps any noisy input $\mathbf{x}_t$ at time $t$ back to a clean sample $\mathbf{x}_{\epsilon}$ at the initial time. This eliminates the need for iterative sampling while maintaining sample fidelity in practice.

This mechanism stands in contrast to traditional diffusion models, which rely on a sequence of conditional generative distributions $p_\theta(\mathbf{x}_{t-1} \mid \mathbf{x}_t)$ to reverse the noise process. By bypassing this iterative process, CM offers a more efficient yet effective alternative for sample generation.

\section{Planning with Consistency Trajectory Model}
\label{sec:plan_consistency}
This paper explores the integration of CTM into the planning architecture in offline RL. In the following, we discuss how we use CTM for the trajectory optimization process. Section \ref{sec:training_process} details the training process of each component, and Section \ref{sec:inference_process} describes how the consistency trajectory planner is applied during inference.

\subsection{Training process}
\label{sec:training_process}
\textbf{Trajectory representation.}
As outlined by \citet{ajay2022conditional}, the diffusion process encompasses only the state transitions, as described by
\begin{align}
    \label{eq:diffuseoverstates}
    \mathbf{x}_{t_{i}}(\tau) := (s_k, s_{k+M}, \ldots, s_{k+(H-1)M})_{t_{i}}.
\end{align}

In this notation, $k$ indicates the timestep of a state within a trajectory $\tau$, $H$ represents the planning horizon, and $t_i \in [\epsilon,t_N]$ is the timestep in the diffusion sequence. To make the model look ahead farther, we choose a jump-step planning strategy \citep{lu2025makes}. Jump-step planning models $H \times M$ environment steps, where $M \in N_+$ is the planning stride. Consequently, $\mathbf{x}_{t_i}(\tau)$ is defined as a noisy sequence of states, represented as a two-dimensional array where each column corresponds to a different timestep of the trajectory. 
In the training process, the sub-sequence $t_i$ follows the Karras boundary schedule \citep{karras2022elucidating}: 
\begin{align}
    \label{eq:training_ti}
    t_i=\left ( \epsilon ^{1/\rho }+\frac{i-1}{N-1}\left ( t_{N}^{1/\rho }  -\epsilon ^{1/\rho} \right )    \right )^{\rho },
\end{align}
where $\epsilon=0.002$, $t_N=80$, and $\rho=7$.

\textbf{Inverse dynamics model training.}
To derive actions from the states generated by the diffusion model, we employ an inverse dynamics model \citep{agrawal2016learning,pathak2018zero}, denoted as $h_{\boldsymbol{\varphi}}$, trained using $(s_k, s_{k+M})$ as input, $a_k$ as target, sampled from the dataset $\mathcal{D}$ consisting of trajectories. Therefore, actions can be obtained via the inverse dynamics model by extracting the state tuple $(s_k, s_{k+M})$ at diffusion timestep $t_0$:

\begin{equation}
    \begin{aligned}
        \label{eq:inv_dy_loss}
        \mathcal{L}({\boldsymbol{\varphi}}):= &\mathbb{E}_{(s_k,a_k,s_{k+M}) \sim \mathcal{D}  }\left[\left \|a_k-h_{\boldsymbol{\varphi}}(s_k,s_{k+M})  \right \|_{2}^{2}\right].
    \end{aligned}
\end{equation}

\textbf{Teacher model training.}
 The teacher model (denoted by $D_{\boldsymbol{\phi}}$) is trained using $s_k$ as condition and $(s_k, s_{k+M}, \ldots, s_{k+(H-1)M})$ as the target output, which is conducted using the following loss:
\begin{equation}
    \begin{aligned}
        \label{eq:edm_loss}
        \mathcal{L}({\boldsymbol{\phi}}):= &\mathbb{E}_{\sigma \sim p_{train},\tau \sim \mathcal{D},n \sim \mathcal{N}(0,{\sigma }^{2}\mathrm {I}    )}\left[ \left \| D_{{\boldsymbol{\phi}} }(\mathbf{x}_{\sigma }(\tau),\sigma)-\mathbf{x}_0(\tau) \right \|_{2}^{2}   \right],
    \end{aligned}
\end{equation}

where $p_{train}$ is a log-normal distribution using the design choice from \citet{karras2022elucidating} and $\sigma$ is the noise level sampled from $p_{train}$. 

The teacher model $D_{\boldsymbol{\phi}}$ is parameterized using a skip-connection architecture, inspired by the formulation in \citet{karras2022elucidating}, and defined as:
\begin{equation}
    \label{eq:D_para}
    D_{\boldsymbol{\phi}}(\mathbf{x}_t, t) = c_{\text{skip}}(t)\, \mathbf{x}_t + c_{\text{out}}(t)\, F_{\boldsymbol{\phi}}(\mathbf{x}_t, t),
\end{equation}
where $F_{\boldsymbol{\phi}}$ is a neural network that outputs the residual signal to correct $\mathbf{x}_t$. The scalar functions $c_{\text{skip}}(t)$ and $c_{\text{out}}(t)$ are time-dependent coefficients that control the contribution of the input and the network output at each timestep $t$. These coefficients are chosen such that $c_{\text{skip}}(\epsilon) = 1$ and $c_{\text{out}}(\epsilon) = 0$, ensuring that the model exactly reproduces the clean data at the boundary $t = \epsilon$, i.e., $D_{\boldsymbol{\phi}}(\mathbf{x}_{\epsilon}, \epsilon) \equiv \mathbf{x}_0$.

This skip and output scaling strategy, inspired by the design proposed in \citet{karras2022elucidating}, allows us to initialize the student neural network $g_{\boldsymbol{\theta}}(\mathbf{x}_t, t, w)$ as a direct copy of the pretrained teacher model $D_{\boldsymbol{\phi}}(\mathbf{x}_t, t)$ to promote fast convergence during training. The only difference lies in the addition of an $w$-embedding module, which enables the student model to condition on both $t$ and $w$.

\textbf{CTM training.} The student model $G_{\boldsymbol{\theta}} $ is expressed as a mixture of $\mathbf{x}_t$ and a neural output $g_{\boldsymbol{\theta}}$:
\begin{align}
    \label{eq:student_nn}
    G_{\boldsymbol{\theta}}\left(\mathbf{x}_t, t, w\right)=\frac{w}{t} \mathbf{x}_t+\left(1-\frac{w}{t}\right) g_{\boldsymbol{\theta}}\left(\mathbf{x}_t, t, w\right).
\end{align}

After the training of the teacher model $D_{\boldsymbol{\phi}}$, 
we distill knowledge from $D_{\boldsymbol{\phi}}$ to train the student model $G_{\boldsymbol{\theta}}$ 
by comparing two predictions at time $w$: one from the student model, $G_{\boldsymbol{\theta}}(\mathbf{x}_t,t,w)$, the other from the target prediction $G_{target}$, using the target model, $G_{\mathrm{sg}{({\boldsymbol{\theta}})}}$, with $\mathrm{sg}({\boldsymbol{\theta}}) \leftarrow (\mu \cdot \mathrm{sg}({\boldsymbol{\theta}})+(1-\mu)\cdot{\boldsymbol{\theta}})$.  More specifically, we randomly sample $u \in [w,t)$ and solve the $(u,t)$-interval using the pretrained probability flow ODE Solver. Then, we jump to time $w$ using the target model to obtain $G_{target}$, which is defined as

\begin{align}
    \label{eq:soft_consis_match}
    G_{target} = G_{\mathrm{sg}({\boldsymbol{\theta}})}(\mathrm{Solver}(\mathbf{x}_t,t,u;{\boldsymbol{\phi}}),u,w).
\end{align}

Here, $\mathrm{Solver}(\mathbf{x}_t,t,u;{\boldsymbol{\phi}})$ is a second-order Heun solver which is applied to solve Eq. \ref{eq:solver} with an initial value of $\mathbf{x}_t$ at time $t$ and ending at time $u$ \citep{karras2022elucidating}:

\begin{equation}
    \label{eq:solver}
    \int_{t}^{u} \frac{d\mathbf{x}_{t'}}{dt'} \, dt' = \int_{t}^{u} \frac{\mathbf{x}_{t'} - D_{\boldsymbol{\phi}}(\mathbf{x}_{t'}, t')}{t'} \, dt' \iff \mathbf{x}_u = \mathbf{x}_t + \int_{t}^{u} \frac{\mathbf{x}_{t'} - D_{\boldsymbol{\phi}}(\mathbf{x}_{t'}, t')}{t'} \, dt'.
\end{equation}

At $u=w$, Eq. \ref{eq:soft_consis_match} enforces global consistency, i.e., the student distills the teacher information on the entire interval $(w,t)$ by comparing $G_{{\boldsymbol{\theta}}}(\mathbf{x}_t,t,w)$ with $\mathrm{Solver}(\mathbf{x}_t,t,w;\phi)$. At $u = t - \Delta t$, Eq. \ref{eq:soft_consis_match} becomes the consistency distillation loss which distills the local consistency, i.e., the student only distills the teacher information on a single-step interval $(t-\Delta t, t)$. In practical implementation, after we obtain these two predictions at time $w$, we transport them to time $0$ using the stop-gradient $G_{\text{sg}}({\boldsymbol{\theta}})(\cdot\, ,\, w, 0)$ and calculate the CTM loss by:

\begin{align}
    \label{eq:ctm_distillation_loss}
    \mathcal{L}_{\mathrm{CTM}}(\boldsymbol{\theta} ; \boldsymbol{\phi}):=\mathbb{E}_{t \in[0, t_N]} \mathbb{E}_{w \in[0, t]} \mathbb{E}_{u \in[w, t)} \mathbb{E}_{\mathbf{x}_0} \mathbb{E}_{\mathbf{x}_t \mid \mathbf{x}_0}\left[d\left(\mathbf{x}_{\text {target }}\left(\mathbf{x}_t, t, u, w\right), \mathbf{x}_{\mathrm{est}}\left(\mathbf{x}_t, t, w\right)\right)\right],
\end{align}

where $d(\cdot , \cdot)$ is the squared $\ell_{2}$ distance with $d(x,y) =\left \|(x-y)  \right \|^2_2$, and $\mathbf{x}_{\mathrm{est}}\left(\mathbf{x}_t, t, w\right):=G_{\mathrm{sg}(\boldsymbol{\theta})}\left(G_{\boldsymbol{\theta}}\left(\mathbf{x}_t, t, w\right), w, 0\right)$, $\mathbf{x}_{\mathrm{target}}\left(\mathbf{x}_t, t, u,w\right):=G_{\mathrm{sg}(\boldsymbol{\theta})}\left(G_{target}, w, 0\right)$, representing the predictions at time $0$. 

To address the gradient vanishing issue that arises when $w\rightarrow t$ in the student model $G_{{\boldsymbol{\theta}}}$---due to the diminishing influence of the neural network component $g_{{\boldsymbol{\theta}}}$ in Eq. \ref{eq:student_nn}---the denoising score matching (DSM) loss is defined as

\begin{align}
    \label{eq:dsm_loss}
    \mathcal{L}_{\mathrm{DSM}}(\boldsymbol{\theta})=\mathbb{E}_{t, \mathbf{x}_0} \mathbb{E}_{\mathbf{x}_t \mid \mathbf{x}_0}\left[\left\|\mathbf{x}_0-g_{\boldsymbol{\theta}}\left(\mathbf{x}_t, t, t\right)\right\|_2^2\right],
\end{align}

which acts as a regularization mechanism that strengthens the learning signal and improve the accuracy when $w \approx t$.

It is noted that in generative tasks, student models obtained through distillation often produce samples of lower quality compared to their teacher counterparts. To address this issue, CTM extends the conventional distillation framework—typically used in classification—by incorporating adversarial learning signals to enhance student training.

Given that GAN-based losses have shown strong performance in image generation tasks \citep{kim2023consistency}, we explore their applicability to trajectory generation. 
The GAN loss is formulated as:
\begin{align}
    \label{eq:gan_loss}
    \mathcal{L}_{\mathrm{GAN}}(\boldsymbol{\theta}, \boldsymbol{\eta}) =
    \mathbb{E}_{\mathbf{x}_0}\left[\log d_{\boldsymbol{\eta}}\left(\mathbf{x}_0\right)\right] +
    \mathbb{E}_{t \in [0, t_N]} \mathbb{E}_{w \in [0, t]} \mathbb{E}_{\mathbf{x}_0} \mathbb{E}_{\mathbf{x}_t \mid \mathbf{x}_0} \left[
    \log \left(1 - d_{\boldsymbol{\eta}}\left(\mathbf{x}_{\mathrm{est}}\left(\mathbf{x}_t, t, w\right)\right)\right)
    \right],
\end{align}

where $d_{\boldsymbol{\eta}}$ is the discriminator function, and $\boldsymbol{\eta}$ denotes its parameters.

In summary, CTM incorporates $\mathcal{L}_{\mathrm{CTM}}$, $\mathcal{L}_{\mathrm{DSM}}$ and $\mathcal{L}_{\mathrm{GAN}}$ as Algorithm \ref{alg:CTM_training} summarizes the training algorithm of CTM:

\begin{align}
    \label{eq:total_loss}
    \mathcal{L}(\boldsymbol{\theta}, \boldsymbol{\eta}):=\mathcal{L}_{\mathrm{CTM}}(\boldsymbol{\theta} ; \boldsymbol{\phi})+\lambda_{\mathrm{DSM}} \mathcal{L}_{\mathrm{DSM}}(\boldsymbol{\theta})+\lambda_{\mathrm{GAN}} \mathcal{L}_{\mathrm{GAN}}(\boldsymbol{\theta}, \boldsymbol{\eta}).
\end{align}

\begin{algorithm}
\caption{Consistency Trajectory Planner's Training}
\label{alg:CTM_training}
\begin{algorithmic}[1]
\Repeat
    \State Sample $\mathbf{x}_{0}(\tau):=(s_k, s_{k+M}, \ldots, s_{k+(H-1)M})$ from data distribution
    \State Sample $\boldsymbol{\epsilon} \sim \mathcal{N}(0, I)$
    \State Sample $t \in [\epsilon, t_N],\ w \in [0, t],\ u \in [w, t)$
    \State Calculate $\mathbf{x}_t = \mathbf{x}_0 + t \boldsymbol{\epsilon}$
    \State Calculate $\textsc{Solver}(\mathbf{x}_t, t, u;\boldsymbol{\phi})$
    \State Update $\boldsymbol{\theta} \gets \boldsymbol{\theta} - \frac{\partial}{\partial \boldsymbol{\theta}} \mathcal{L}(\boldsymbol{\theta}, \boldsymbol{\eta})$
    \State Update $\boldsymbol{\eta} \gets \boldsymbol{\eta} + \frac{\partial}{\partial \boldsymbol{\eta}} \mathcal{L}_{\text{GAN}}(\boldsymbol{\theta}, \boldsymbol{\eta})$
\Until{converged}
\end{algorithmic}
\end{algorithm}

\textbf{Critic Model training.}
During the sampling process, we use Monte Carlo sampling from selections, where $N$ selections are firstly sampled from CTM as candidates. To select the best plan from the $N$ selections, a critic model $V_{\boldsymbol{\alpha}}$ is trained using $\mathbf{x}_{t_{0}}(\tau)$ as input, the accumulated discounted returns $R_k$ as target output, where ${\boldsymbol{\alpha}}$ denotes the parameters of the critic network. Specifically, $R_k$ is calculated by
\begin{align}
    \label{eq:culmulated_r}
    R_k = \sum_{h=0}^{t_{end}} \gamma^h r_{k+h}.
\end{align}

To train the critic model $V_{\boldsymbol{\alpha}}$, we minimize the mean squared error between the predicted return and the actual accumulated return:
\begin{align}
    \label{eq:critic_loss}
    \mathcal{L}_{\text{critic}}({\boldsymbol{\alpha}}) = \mathbb{E}_{\tau \sim \mathcal{D}} \left[ \left( V_{\boldsymbol{\alpha}}(\mathbf{x}_{t_0}(\tau)) - R_k \right)^2 \right].
\end{align}

This value-based selection process enables the planner to rank candidate trajectories according to their expected returns and choose the highest-scoring plan for execution.

\subsection{Inference process}
\label{sec:inference_process}
\textbf{Sampling with CTM.}
To generate samples using CTM, we follow a reverse-time denoising procedure along a predefined sequence of time steps ${t_0, t_1, \dots, t_N}$, where $t_n = \frac{n}{N}{t_N}$ and $t_0 = \epsilon$. Sampling begins by drawing an initial noisy sample $\mathbf{x}_{t_N}$ from a standard Gaussian prior $\mathcal{N}(\mathbf{0}, t_N^2 \mathbf{I})$. Then, for each time step $n = N-1, \dots, 0$, the model applies a single-step denoising operation using the student model $G_{\boldsymbol{\theta}}$, which maps the current noisy sample $\mathbf{x}_{t_{n+1}}$ from time $t_{n+1}$ to a less noisy representation at time $t_n$. 

Formally, this process iteratively computes
\begin{align}
    \label{eq:student_model_sampling}
    \mathbf{x}_{t_{n}} \gets G_{\boldsymbol{\theta}}(\mathbf{x}_{t_{n+1}}, t_{n+1}, {t_{n}}),
\end{align}

until reaching the final output $\mathbf{x}_{t_0}$, which serves as the generated sample. Unlike the inference process in CM, we do not perturb the neural sample with noise to avoid error accumulation. 
This deterministic sampling procedure allows CTM to produce high-quality samples in a small number of steps, while preserving flexibility in the choice of discretization schedule.

\begin{figure}[t]
\begin{center}
\includegraphics[width=\textwidth, trim=0cm 8cm 0cm 13cm, clip]{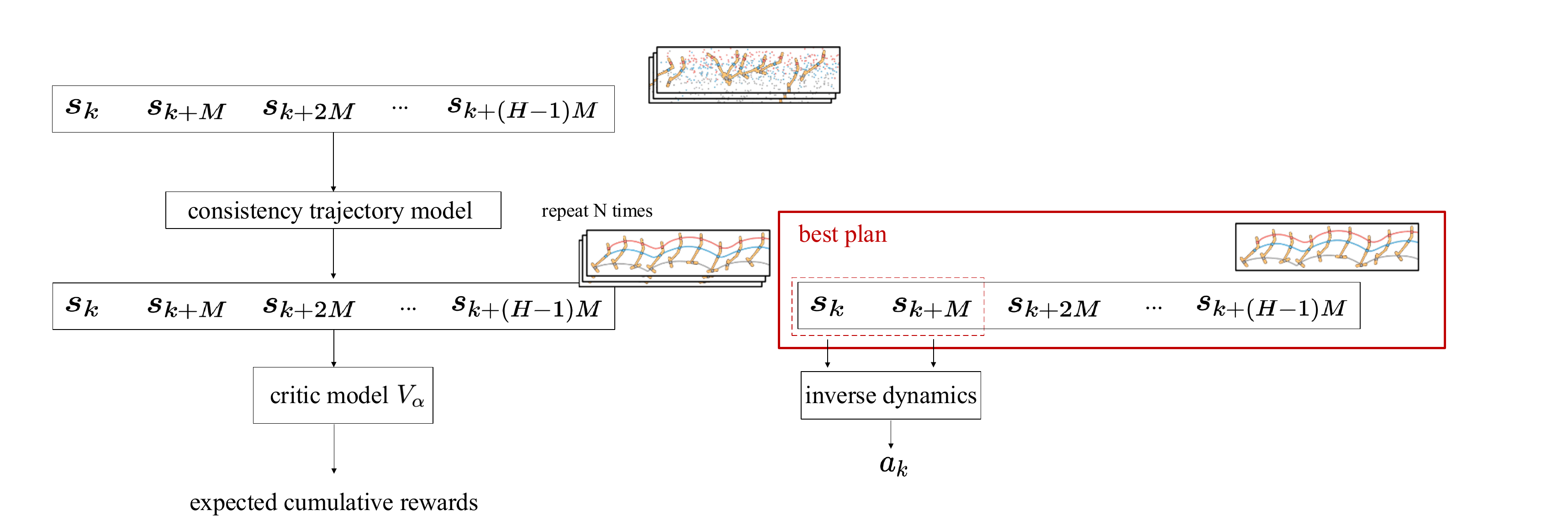}
\end{center}
\caption{Consistency Trajectory Planning. Given the current state $s_k$, Consistency Trajectory Planning generates $N$ sequences of future states with planning horizon $H$. Then, the best plan is selected by the critic model. Finally, the inverse dynamics model is used to extract and execute the action $a_k$ from $s_k$ and $s_{k+M}$ which are from the selected best plan.}
\label{fig:framework}
\end{figure}

\begin{algorithm}
\caption{Consistency Trajectory Planning}
\label{alg:ctp_framework}
\begin{algorithmic}[1] 
\State \textbf{Input}: Planning horizon $H$, Dataset $\mathcal{D}$, Discount factor $\gamma$, Candidate num $N$, Planning stride $M$
\State \textbf{Initialize}: Diffusion Transformer Planner $D_{\boldsymbol{\phi}}$, Consistency Trajectory Planner $g_{\boldsymbol{\theta}}$, Diffusion Inverse dynamics $h_{\boldsymbol{\varphi}}$, Critic $V_{\boldsymbol{\alpha}}$
\State Calculate accumulated discounted returns $R_k = \sum_{h=0}^{\text{end}} \gamma^h r_{k+h}$ for every step $k$
\Function{Training}{}
    \State Sample $s_k, s_{k+M}, \dots, s_{k+(H-1)M}$, $a_k,a_{k+M},\dots,a_{k+(H-1)M}$, $R_k$ from $\mathcal{D}$
    \State Train diffusion model $D_{\boldsymbol{\phi}}$ using $s_k$ as condition and $s_{k,k+M,\dots,k+(H-1)M}$ as target output (Eq. \ref{eq:edm_loss})
    \State Distill consistency trajectory planner $g_{\boldsymbol{\theta}}$ (Eq. \ref{eq:total_loss})
    \State Train inverse dynamics $h_{\boldsymbol{\varphi}}$ using $s_k, s_{k+M}$ as input, $a_k$ as target output (Eq. \ref{eq:inv_dy_loss})
    \State Train critic $V_{\boldsymbol{\alpha}}$ using $s_{k, k+M, \dots, k+(H-1)M}$ as input, $R_k$ as target output (Eq. \ref{eq:critic_loss})
\EndFunction
\Function{Planning}{$s$}
    \State Randomly generate $N$ plans using CTP sampling, while fixing the first state as $s$ during sampling
    \State Select the best plan using critic $V_{\boldsymbol{\alpha}}$
    \State Use the inverse dynamics $h_{\boldsymbol{\varphi}}$ to generate action using $s$ and the next state in the best plan
\EndFunction
\end{algorithmic}
\end{algorithm}

\textbf{CTP Inference.}
During the inference process, we first observe a state $s$ in the environment and sample a Gaussian noise $\mathbf{x}_{t_N}$. Then, CTM iteratively predicts the denoised trajectories $\mathbf{x}_{t_n}$ from the noisy inputs. This process is repeated $N$ times to generate $N$ plans using $g_{{\boldsymbol{\theta}}}$, while the first state $s$ is fixed during sampling. Then, we select the best plan using critic $V_{\alpha}$. Finally, we extract states $(s_k, s_{k+M})$ from the denoised trajectory and get the action $a_k$ via our inverse dynamics model $h_{\phi}$. 
The algorithm of CTP is provided in Algorithm \ref{alg:ctp_framework} and visualized in Figure \ref{fig:framework}.

\section{Experiment}
\label{sec:experiment}
In this section, we present the experiment environment, experiment setting and report empirical results that validate the effectiveness of the proposed CTP algorithm across diverse offline RL tasks.

\subsection{Experiment Environment}
\label{sec:exp_environment}

We train the diffusion model, inverse dynamics model, and consistency trajectory model on publicly available D4RL datasets. Evaluation is conducted across diverse Gym tasks, including locomotion (HalfCheetah, Hopper, Walker2d), long-horizon planning (Maze2D), and high-dimensional robotic control tasks (Antmaze, Kitchen, Adroit) from the D4RL benchmark suite \citep{fu2020d4rl}.
These tasks are characterized by continuous state and action spaces and are conducted under offline RL settings. Details of the dataset size are provided in Appendix.

\textbf{Locomotion}
The locomotion tasks—Hopper, HalfCheetah, and Walker2d—are widely adopted due to their controlled dynamics, reproducibility, and varying levels of complexity. These environments are based on the MuJoCo physics engine \citep{todorov2012mujoco} and simulate planar bipedal or quadrupedal agents that must learn to move forward efficiently.
\begin{itemize}
\item    \textbf{Hopper} involves a single-legged robot that must learn to hop forward without falling, presenting challenges in stability and balance.
\item    \textbf{HalfCheetah} features a planar cheetah-like agent with a more complex morphology, requiring the agent to coordinate multiple joints to achieve high-speed locomotion.
\item    \textbf{Walker2d} simulates a two-legged humanoid robot that must walk forward while maintaining upright posture, making it more prone to instability.
\end{itemize}
These tasks serve as standard benchmarks for evaluating the ability of offline algorithms to generalize from limited data and produce smooth, efficient motion without direct online interaction.

\textbf{Maze2D}
To validate the long-horizon planning capabilities of CTP, we conduct an evaluation in the Maze2D environment \citep{fu2020d4rl}, where the task involves navigating to a specific goal location, with a reward of $1$ assigned only upon reaching the goal. Because it requires hundreds of steps to reach the goal, even the most advanced model-free algorithms struggle with effective credit assignment and consistently reaching the goal.

\textbf{AntMaze}
The AntMaze task extends the Maze2D environment by replacing the simple 2D ball agent with a more complex quadrupedal "Ant" robot, thereby combining challenges of both locomotion and high-level planning. In the diverse variant of the dataset, the ant is initialized at random positions and directed toward randomly sampled goals. In contrast, the play variant consists of trajectories guided toward manually selected goal locations within the maze.

\textbf{Kitchen}
The Kitchen environment simulates a robotic manipulator interacting with various appliances in a realistic kitchen setting, requiring the agent to perform multi-stage manipulation tasks to accomplish specified goals. In the partial dataset, only a subset of trajectories successfully achieves the full task, allowing imitation-based methods to benefit from selectively identifying informative demonstrations. The mixed dataset, on the other hand, contains no trajectories that solve the task in its entirety, necessitating the use of RL to stitch together relevant sub-trajectories.

\textbf{Adroit}
The Adroit Hand benchmark consists of four high–degree-of-freedom manipulation tasks constructed from motion-capture demonstrations of human hand movements \citep{rajeswaran2017learning, fu2020d4rl}. The benchmark couples challenging planning objectives with low-level motor control, thereby providing a stringent test of our approach.

The benchmark comprises four distinct manipulation scenarios, each instantiated with a 30-DoF dexterous hand mounted on a freely moving arm:
\begin{itemize}

\item    \textbf{Door.} The agent must disengage a latch with substantial dry friction and swing the door until it contacts the stopper. No explicit latch state is provided; the agent infers its dynamics solely through interaction. The door’s initial pose is randomized across trials.

\item    \textbf{Hammer.} The hand picks up a hammer and drives a nail of variable location into a board. The nail—subject to dry friction resisting up to 15N—must be fully embedded for success.

\item    \textbf{Pen.} With the wrist fixed, the agent reorients a blue pen so that its pose matches a randomly placed green target within a prescribed angular tolerance.

\item    \textbf{Relocate.} The agent transports a blue sphere to a green target whose position, along with the sphere’s start pose, is uniformly randomized throughout the workspace; success is declared once the sphere lies within an $\epsilon$-ball of the target.
\end{itemize}

\subsection{Experiment Setting}
\label{sec:setting}

\begin{table}[t]
\caption{The average scores of Diffuser, Decision Diffuser, Diffusion-QL, Consistency-AC, Consistency Planning and our method on D4RL locomotion tasks are shown. The results of previous work are quoted from \citet{ding2023consistency}, \citet{ajay2022conditional} and \citet{wangplanning}.}
    \label{tab:Table_locomotion}
    \begin{center}
        \begin{tabular}{llcccccc}
            \multicolumn{1}{l}{\bf Dataset} &\multicolumn{1}{l}{\bf Environment}  &\multicolumn{1}{l}{\bf Diffuser} &\multicolumn{1}{l}{\bf DD} &\multicolumn{1}{l}{\bf D-QL} &\multicolumn{1}{l}{\bf C-AC} &\multicolumn{1}{l}{\bf CP} &\multicolumn{1}{l}{\bf CTP}
            \\ \hline \\
            Medium-Expert &Halfcheetah &79.8&90.6&\textbf{96.8}&84.3&94&$89.3\pm0.6$ \\
            Medium-Replay &Halfcheetah &42.2&39.3&47.8&\textbf{58.7}&40.6&$43.4\pm0.4$ \\ 
            Medium &Halfcheetah &44.2&49.1&51.1&\textbf{69.1}&46.8&$50.4\pm0.1$ \\

            Medium-Expert &Hopper& 107.2&\textbf{111.8}&111.1&100.4&107.5&$107.5\pm1.2$\\
            Medium-Replay &Hopper&96.8&100&\textbf{101.3}&99.7&97.8&$90.0\pm1.0$ \\
            Medium &Hopper&58.5&79.3&\textbf{90.5}&80.7&87.8&$83.6\pm1.3$\\
            
            Medium-Expert &Walker2d& 108.4&108.8&110.1&\textbf{110.4}&109.8& \textbf{$110.1\pm0.05$}\\
            Medium-Replay &Walker2d&61.2&75&\textbf{95.5}&79.5&75.3&$86.9\pm0.3$ \\
            Medium &Walker2d&79.7&82.5&\textbf{87.0}&83.1&80.5&$85.7\pm0.2$\\
            \\ \hline \\
            \textbf{Average}&-&75.3&81.8&\textbf{87.9}&85.1&82.2&$83$\\

        \end{tabular}
    \end{center}
    
\end{table}

We evaluate the performance of our proposed method by comparing it against both actor-critic-based approaches, including Consistency Actor-Critic (C-AC) \citep{ding2023consistency} and Diffusion-QL (D-QL) \citep{wang2022diffusion}, as well as model-based planning methods, such as Diffuser \citep{janner2022planning}, Decision Diffuser (DD) \citep{ajay2022conditional}, Consistency Planning (CP) \citep{wangplanning}, Reward-Aware Consistency Trajectory Distillation (RACTD) \citep{duan2025accelerating} and Lower Expectile Q-learning (LEQ) \citep{parkmodel}.

For all experiments, we report results as the mean over 150 independent planning seeds to ensure statistical robustness. Following the evaluation protocol established in \citet{fu2020d4rl}, we adopt the normalized average return as the primary performance metric.

Unless otherwise specified, the CTP algorithm employs $N=2$ denoising steps, which we found to yield near-saturated performance across most tasks. An exception is the Maze2D domain, where we use a single denoising step ($N=1$) due to the relative simplicity of the environment. For comparison, competing methods employ different denoising schedules: the diffusion policy utilizes $N=5$ steps \citep{wang2022diffusion}, Diffuser adopts $N=20$ steps, and Decision Diffuser uses $N=40$ steps \citep{janner2022planning, ajay2022conditional}. Consistency-based baselines such as Consistency Actor-Critic and Consistency Planning are evaluated with $N=2$ denoising steps, consistent with their original implementations.

\begin{table}[t]
\caption{The performance of CTP, Diffuser, and previous model-free algorithms in the Maze2D environment, which tests long-horizon planning due to its sparse reward structure. The results of previous work are quoted from the data provided in \citet{janner2022planning}, \citet{wangplanning} and \citet{duan2025accelerating}.}
    \label{tab:Table_maze2d}
    \begin{center}
        \begin{tabular}{lccccccc}
            \multicolumn{1}{l}{\bf Dataset} &\multicolumn{1}{l}{\bf MPPI} &\multicolumn{1}{l}{\bf CQL} &\multicolumn{1}{l}{\bf IQL} &\multicolumn{1}{l}{\bf Diffuser} &\multicolumn{1}{l}{\bf CP} &\multicolumn{1}{l}{\bf RACTD}&\multicolumn{1}{l}{\bf CTP}
            \\ \hline \\
            Maze2D U-Maze&33.2&5.7&47.4&$113.9 \pm 3.1$& 122.7$ \pm 2.7$ & $125.7 \pm 0.6$& $\textbf{154.1} \pm 2.3$ \\
            Maze2D Medium&10.2&5.0&34.9&121.5$ \pm 2.7$ & 121.4$\pm 4.1$& $130.8 \pm 1.8$&$\textbf{167.1}\pm2.4$  \\
            Maze2D Large&5.1&12.5&58.6&123.0$ \pm 6.4$ & $119.5 \pm 7.5$& $143.8 \pm 0.0$&$\textbf{216.7} \pm 3.4$
            \\ \hline \\
            Average&16.2&7.7&47.0&$119.5$&121.2&133.4&\textbf{179.3}
        \end{tabular}
    \end{center}
    
\end{table}

\begin{table}[t]
\caption{The performance of CTP, Diffuser, and previous model-free algorithms in the Kitchen environment, which tests  both locomotion and high-level planning capability. The results of previous work are derived from the data provided in \citet{ding2023consistency} and \citet{ajay2022conditional}.}
    \label{tab:kitchen}
    \begin{center}
        \begin{tabular}{llccccccc}
            \multicolumn{1}{l}{\bf Dataset} &\multicolumn{1}{l}{\bf Environment}  &\multicolumn{1}{l}{\bf Diffuser} &\multicolumn{1}{l}{\bf DD} &\multicolumn{1}{l}{\bf D-QL} &\multicolumn{1}{l}{\bf C-AC} &\multicolumn{1}{l}{\bf CTP}
            \\ \hline \\
            Mixed &Kitchen &47.5 &$65\pm2.8$  &$62.6\pm5.1$&$45.8\pm1.5$ &$\textbf{74.5}\pm0.3$ \\
            Partial &Kitchen &33.8 &$57\pm2.5$ &$60.5\pm6.9$ &$38.2\pm1.8$&$\textbf{91.2}\pm1.0$ \\
            
            \\ \hline \\
            \textbf{Average}&-&40.65&61&61.55&42&\textbf{82.85}\\

        \end{tabular}
    \end{center}
    
\end{table}

\begin{table}[t]
\caption{The performance of CTP and LEQ in the Antmaze environment. The results of LEQ are quoted from \citet{parkmodel}.}
    \label{tab:antmaze}
    \begin{center}
        \begin{tabular}{llll}
            \multicolumn{1}{l}{\bf Dataset} &\multicolumn{1}{l}{\bf Environment} &\multicolumn{1}{l}{\bf LEQ} &\multicolumn{1}{l}{\bf CTP}
            
            \\ \hline \\
            Diverse& Antmaze-Large  &$60.2\pm18.3$ &$\textbf{82.0}\pm3.1$  \\
            Play&Antmaze-Large &$62.0\pm9.9$ &$\textbf{82.0}\pm3.1$ \\
            Diverse&Antmaze-Medium  &$46.2\pm23.2$ &$\textbf{86.0}\pm2.8$  \\
            Play&Antmaze-Medium  &$76.3\pm17.2$ &$\textbf{83.3}\pm3.0$  \\
            \\ \hline \\
            \textbf{Average}&-&61.18&\textbf{83.33}\\
        
        \end{tabular}
    \end{center}
    
\end{table}

\begin{table}[t]
\caption{The performance of CTP against other baselines in the Adroit environment. The results of previous work are quoted from the data provided in \citet{he2024aligniql}.}
    \label{tab:adroit}
    \begin{center}
        \begin{tabular}{llcccccc}
            \multicolumn{1}{l}{\bf Dataset} &\multicolumn{1}{l}{\bf Environment} &\multicolumn{1}{l}{\bf BC} &\multicolumn{1}{l}{\bf BCQ} &\multicolumn{1}{l}{\bf CQL} &\multicolumn{1}{l}{\bf IQL} &\multicolumn{1}{l}{\bf AlignIQL} &\multicolumn{1}{l}{\bf CTP}
            
            \\ \hline \\
            Expert& door &34.9 &99.0 &101.5 &103.8 &\textbf{104.6} &$\textbf{104.1}\pm0.5$  \\
            Expert& hammer &\textbf{125.6} &107.2 &86.7 &116.3 &\textbf{124.7} &$110.5\pm3.1$ \\
            Expert& pen &85.1 &114.9 &107.0 &111.7 &\textbf{116.0} &$104.1\pm5.3$  \\
            Expert& relocate &101.3 &41.6 &95.0 &102.7 &106.0 &$\textbf{108.8}\pm0.6$  \\
            \\ \hline \\
            \textbf{Average}&- &86.7 &90.7 &97.6 &108.6 &\textbf{112.8} &106.88\\
        
        \end{tabular}
    \end{center}
    
\end{table}

\begin{figure}[t]
 \centering
 \begin{subfigure}[b]{0.32\textwidth}
   \centering
   \includegraphics[width=\linewidth]{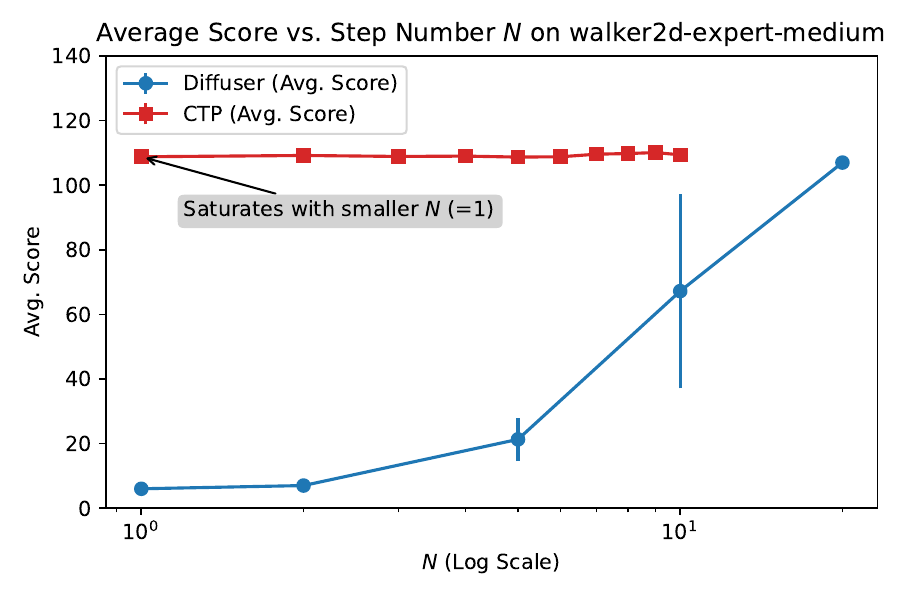}
   \caption{}
   \label{fig:score_wme]]}
 \end{subfigure}
 \hfill
 \begin{subfigure}[b]{0.32\textwidth}
   \centering
   \includegraphics[width=\linewidth]{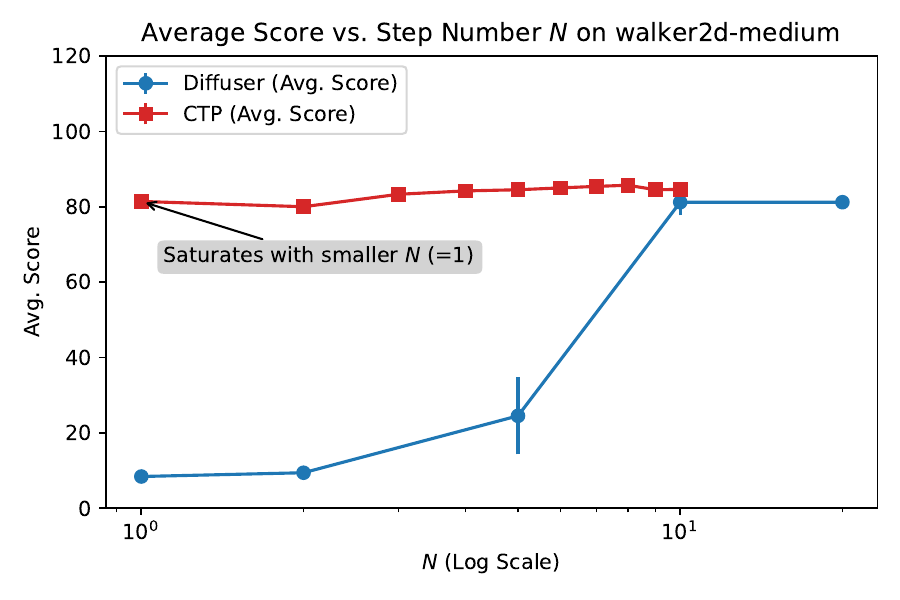}
   \caption{}
   \label{fig:score_wm}
 \end{subfigure}
 \hfill
 \begin{subfigure}[b]{0.32\textwidth}
   \centering
   \includegraphics[width=\linewidth]{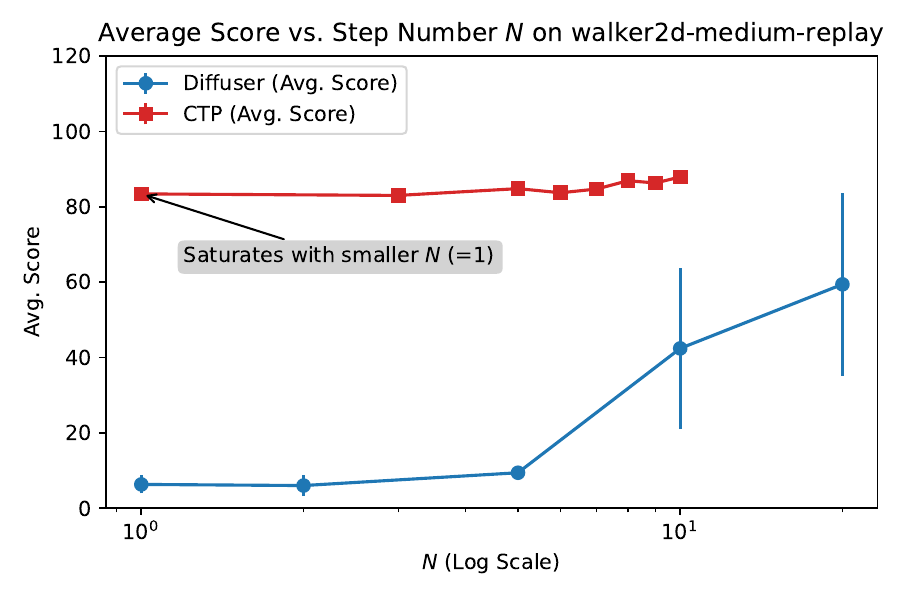}
   \caption{}
   \label{fig:score_wmr}
 \end{subfigure}

 \vspace{0.8em} 

 \begin{subfigure}[b]{0.32\textwidth}
   \centering
   \includegraphics[width=\linewidth]{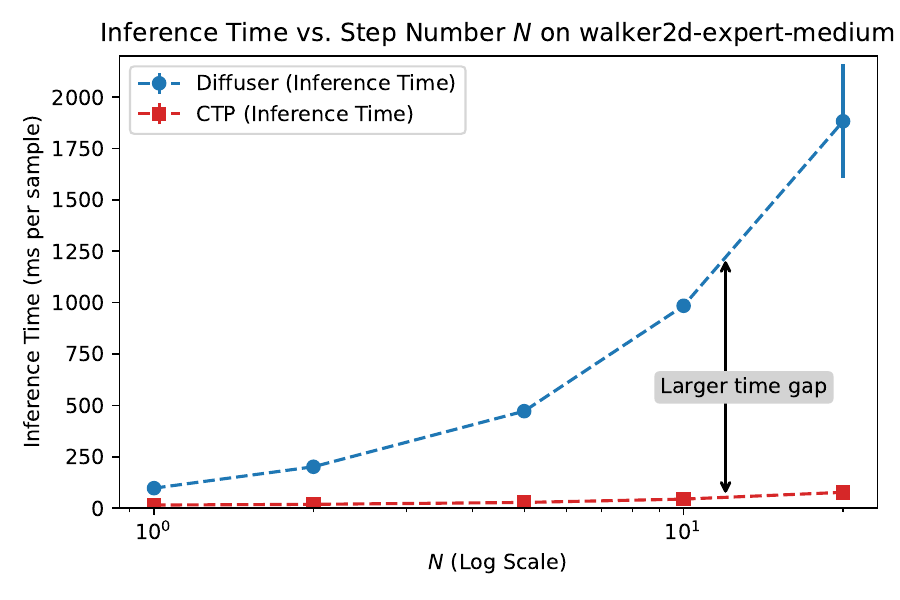}
   \caption{}
   \label{fig:time_wme}
 \end{subfigure}
 \hfill
 \begin{subfigure}[b]{0.32\textwidth}
   \centering
   \includegraphics[width=\linewidth]{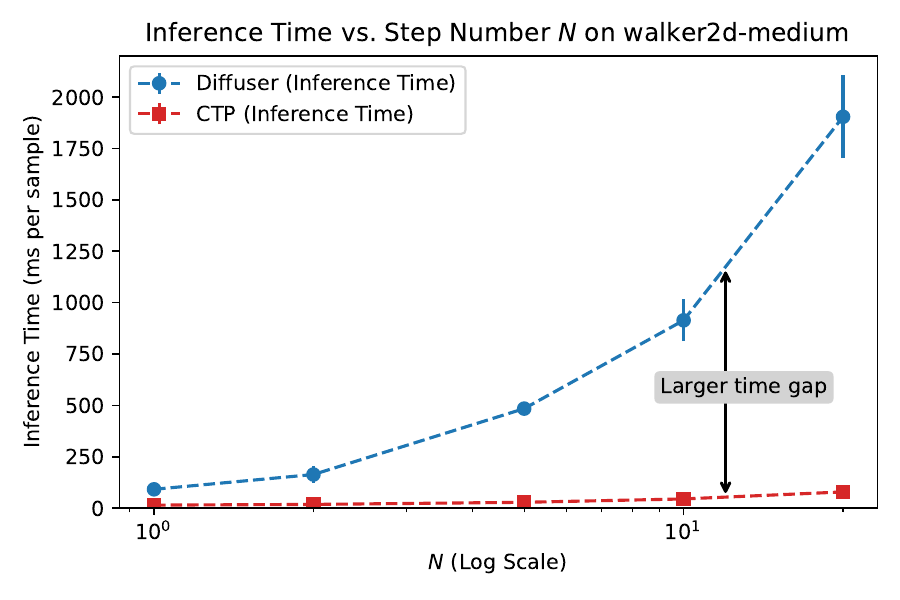}
   \caption{}
   \label{fig:time_wm}
 \end{subfigure}
 \hfill
 \begin{subfigure}[b]{0.32\textwidth}
   \centering
   \includegraphics[width=\linewidth]{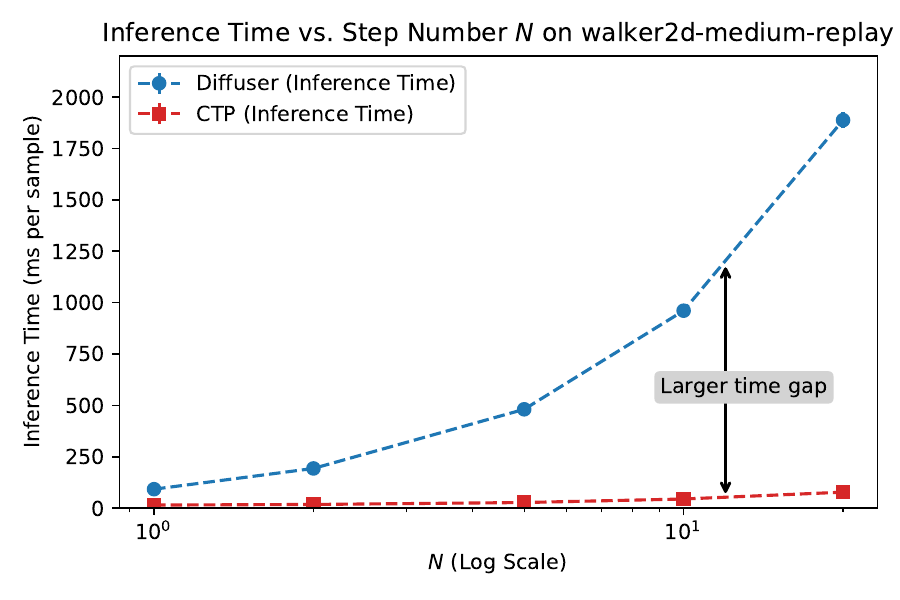}
   \caption{}
   \label{fig:time_wmr}
 \end{subfigure}

 \caption{Comparison of the average normalized scores (top row) and inference times (bottom row) of CTP and Diffuser on three walker2d datasets: walker2d-expert-medium, walker2d-medium, and walker2d-medium-replay. Each column corresponds to a specific dataset. The x-axis denotes the number of denoising steps $N$ on a logarithmic scale. Vertical error bars represent the standard deviation over five random seeds. While CTP achieves near-optimal performance with significantly fewer denoising steps (e.g., saturating at $N=1$), Diffuser requires substantially more steps to match similar scores. Inference time grows rapidly with larger $N$ for Diffuser, whereas CTP maintains consistently low inference time.}
 \label{fig:inference_time}
\end{figure}

\subsection{Experimental Results}
\label{sec:results}

As shown in Table~\ref{tab:Table_locomotion}, CTP achieves competitive performance across standard locomotion benchmarks. These results highlight the effectiveness of CTP in matching the performance of prior state-of-the-art diffusion-based planning algorithms.

Given the relatively lower complexity of Maze2D compared to AntMaze and Kitchen, our model is capable of generating effective trajectories with one-step generation ($N=1$), while still achieving best performance (Table \ref{tab:Table_maze2d}). 
Notably, when compared to the recent concurrent method RACTD~\citep{duan2025accelerating}, which adopts a model-free approach, CTP consistently achieves higher returns across all Maze2D tasks, which require accurate trajectory optimization and modeling of long-term dependencies. 
By leveraging environment dynamics through CTM, CTP enables high-quality planning under sparse-reward, long-horizon scenarios.

Tables \ref{tab:kitchen} and \ref{tab:antmaze} present the performance of CTP on the Kitchen and AntMaze benchmarks, illustrating its robustness in addressing complex, goal-conditioned control tasks. Our experimental results demonstrate that CTP performs competitively not only in relatively simple environments such as Maze2D—where no robotic actuation is involved—but also in more challenging domains like Kitchen and AntMaze, which demand high-dimensional, temporally coordinated control.

We further evaluate CTP on the “expert” dataset from the Adroit benchmark, which consists of $5000$ trajectories generated by a policy that consistently completes the task. Quantitative results are reported in Table~\ref{tab:adroit}. 
While CTP does not outperform all prior methods across every task, it achieves competitive results overall. These findings suggest that CTP is effective in goal-conditioned manipulation scenarios that require precise spatial reasoning and long-horizon planning.

Notably, we observe that diffusion-based policy methods tend to achieve superior performance on MuJoCo locomotion tasks. This can be attributed to the fundamental differences in task structure: environments like AntMaze, Kitchen, and Maze2D require precise goal-directed behavior and long-horizon planning (e.g., manipulating objects into exact positions or reaching distant targets), which aligns well with trajectory-level planning methods such as CTP that generate entire action sequences in a single step. Moreover, these domains are characterized by sparse and delayed rewards, posing significant challenges to model-free RL methods commonly used in diffusion policies \citep{wang2022diffusion, ding2023consistency}.

By contrast, the objective in MuJoCo environments is relatively straightforward—typically maximizing forward velocity—which does not necessitate explicit planning or long-term credit assignment. In such settings, model-free actor-critic approaches, aided by RL loss functions, can more effectively optimize behavior and achieve strong performance \citep{wang2022diffusion, ding2023consistency}.

\textbf{Computational Time.} To assess the computational efficiency of CTP relative to Diffuser under varying denoising step numbers $N$, we conduct a systematic evaluation of both inference time and policy performance on the walker2d-medium-exper task. Since both CTP and Diffuser are built upon generative modeling frameworks based on probability flow, their inference cost scales with the number of denoising steps $N$. However, by design, the CTP enables effective single-step sampling, in contrast to diffusion models which typically require iterative refinement to achieve comparable sample quality.

Figure \ref{fig:inference_time} presents a detailed comparison of the average normalized scores and inference time (in milliseconds per sample) across varying denoising steps. Each data point reports the mean and standard deviation computed over five random seeds. As shown, CTP achieves near-optimal performance with $N=1$, and saturates fully by $N=2$. In contrast, Diffuser requires up to $N=20$ steps to reach its performance plateau, with inference time increasing significantly with each additional step.

Furthermore, Diffuser with $N=20$ requires approximately 1900 ms per sample for inference, while CTP with $N=1$ achieves the same in just 15 ms—resulting in a speedup of roughly $120\times$ under this setting.
Despite this large discrepancy in computation time, CTP continues to outperform Diffuser in terms of policy quality, yielding a slight improvement in normalized score.

These results underscore the efficiency advantage of CTP, demonstrating that it achieves strong performance with substantially fewer denoising steps, thereby offering a more practical solution for time-sensitive deployment scenarios.

\section{Conclusion}
\label{sec:conclusion}

In this work, we propose CTP, a novel approach that integrates CTM into the trajectory optimization framework for offline RL. CTP enables highly effective one-step trajectory sampling, achieving superior performance across standard benchmarks. Our method demonstrates robust generalization in both simple and complex environments, showcasing its practical value for efficient policy planning.

Looking ahead, several avenues of improvement remain. First, enhancing the critic model \citep{kostrikov2021offline} to better estimate the upper bound of plausible returns—rather than the mean—can improve planning guidance; this can be achieved using asymmetric loss functions such as the expectile loss. Second, the consistency loss weighting scheme can be refined to more accurately reflect the impact of different noise scales during training \citep{song2023improved}, thereby amplifying the signal from informative samples. Finally, incorporating advanced network architectures (e.g., UNet, Transformer) along with curriculum learning strategies may further improve training stability and policy performance. We leave these directions for future work.



\bibliographystyle{tmlr}
\newpage
\appendix
\section{Appendix}

Table \ref{tab:dataset_size} provides a comprehensive summary of dataset sizes used in our experiments. All datasets are sourced from the latest release of the D4RL benchmark suite \citep{fu2020d4rl}, encompassing four domains: Gym-MuJoCo, Maze2D, FrankaKitchen, and AntMaze.

In the Gym-MuJoCo domain, we consider three standard environments—Hopper, HalfCheetah, and Walker—each associated with three data variants: medium-expert, medium, and medium-replay.
The medium-expert datasets comprise trajectories generated by a mixture of medium- and expert-level policies, thus incorporating both suboptimal and near-optimal actions. The medium datasets are collected exclusively using medium-performance policies, and consequently, contain a higher proportion of suboptimal behavior. The medium-replay datasets represent replay buffers collected from medium-level agents during training, featuring highly diverse suboptimal trajectories and exploration noise.

In the Maze2D domain, datasets are categorized according to maze complexity: umaze, medium, and large. Each variant reflects increasing navigation difficulty and state space dimensionality.

The FrankaKitchen domain includes two types of datasets: partial and mixed, both of which consist of undirected demonstrations. In the partial dataset, a subset of trajectories successfully accomplish the full task, enabling imitation learning algorithms to leverage these informative samples. Conversely, the mixed dataset contains only partial trajectories, with no complete demonstrations of the full task. This requires RL algorithms to generalize from sub-trajectories and effectively compose them into successful task completions.

For the AntMaze domain, we use the same three maze configurations (umaze, medium, and large) as in Maze2D. Three types of datasets are constructed: In the standard setting (antmaze-umaze-v0), the ant is instructed to reach a fixed goal from a fixed start state. In the diverse datasets, both the goal and starting positions are randomly sampled, introducing high variability. In the play datasets, the ant is commanded to reach hand-picked waypoints in the maze, which may not coincide with the evaluation goal, and also begins from a curated set of start locations. These variations are designed to assess the agent’s ability to generalize under different levels of distributional shift and task ambiguity.

\begin{table}[h]
\centering
\begin{tabular}{llc}
\toprule
\textbf{Domain} & \textbf{Task Name} & \textbf{Samples} \\
\midrule
\multirow{9}{*}{Gym-MuJoCo} 
    & hopper-me & $2 \times 10^6$ \\
    & hopper-m& $10^6$ \\
    & hopper-mr& $402000$ \\
    & halfcheetah-me& $2 \times 10^6$ \\
    & halfcheetah-m& $10^6$ \\
    & halfcheetah-mr& $202000$ \\
    & walker-me& $2 \times 10^6$ \\
    & walker-m& $10^6$ \\
    & walker-mr& $302000$ \\
\midrule
\multirow{3}{*}{Maze2D} 
    & maze2d-umaze       & $10^6$       \\
    & maze2d-medium      & $2 \times 10^6$ \\
    & maze2d-large       & $4 \times 10^6$ \\
\midrule
\multirow{2}{*}{FrankaKitchen}
    & kitchen-mixed &$136950$\\
    & kitchen-partial &$136950$\\
\midrule
\multirow{6}{*}{AntMaze} 
    & antmaze-medium-play     & $10^6$ \\
    & antmaze-medium-diverse  & $10^6$ \\
    & antmaze-large-play      & $10^6$ \\
    & antmaze-large-diverse   & $10^6$ \\
\bottomrule
\end{tabular}
\caption{Size for each dataset is provided. The number of samples indicates the total count of environment transitions recorded in the dataset \citep{fu2020d4rl}.}
\label{tab:dataset_size}
\end{table}

In the next section, we describe various architectural and hyper-parameter details:

\begin{itemize}
    \item We parameterize the diffusion model and consistency trajectory model with a Transformer architecture.
    \item We use Transformer depth of 2 in all the MuJoCo tasks, Kitchen tasks, Maze2D tasks, and Adroit tasks, 8 in AntMaze-Medium tasks, 12 in AntMaze-Large tasks.
    
    \item We train EDM (teacher model) using learning rate of $2e-4$ and CTM (student model) using learning rate of $8e-6$. Both use batch size of $128$ for $1e6$ train steps with Adam optimizer.
    \item We train both inverse dynamics model and critic model using learning rate of $3e-4$.    
    \item We use $N = 2$ for CTP inference.
    \item We use a planning horizon $H$ of 4 in all the MuJoCo tasks, 32 in Kitchen tasks, 32 in Maze2D tasks, 40 in AntMaze tasks, 32 in Adroit tasks.
    \item We use a planning stride $M$ of 1 in all the MuJoCo tasks, 4 in Kitchen tasks, 15 in Maze2D tasks, 25 in AntMaze tasks, 2 in Adroit tasks.
\end{itemize}

\end{document}